\documentclass[letterpaper]{article} % DO NOT CHANGE THIS
\usepackage{aaai25}  % DO NOT CHANGE THIS
\usepackage{times}  % DO NOT CHANGE THIS
\usepackage{helvet}  % DO NOT CHANGE THIS
\usepackage{courier}  % DO NOT CHANGE THIS
\usepackage[hyphens]{url}  % DO NOT CHANGE THIS
\usepackage{graphicx} % DO NOT CHANGE THIS
\usepackage{natbib}  % DO NOT CHANGE THIS AND DO NOT ADD ANY OPTIONS TO IT
\usepackage{caption} % DO NOT CHANGE THIS AND DO NOT ADD ANY OPTIONS TO IT
\usepackage{algorithm}
\usepackage{algorithmic}
\usepackage{subfig}
\usepackage{amsmath}
\usepackage{amssymb}
\usepackage{booktabs}
\usepackage{multirow}

\usepackage{newfloat}
\usepackage{listings}
\DeclareCaptionStyle{ruled}{labelfont=normalfont,labelsep=colon,strut=off} % DO NOT CHANGE THIS
\floatstyle{ruled}
\newfloat{listing}{tb}{lst}{}
\floatname{listing}{Listing}
\title{Predicting the Original Appearance of Damaged Historical Documents}
\author{
    Zhenhua Yang\textsuperscript{\rm 1}\equalcontrib,
    Dezhi Peng\textsuperscript{\rm 1}\equalcontrib,
    Yongxin Shi\textsuperscript{\rm 1},
    Yuyi Zhang\textsuperscript{\rm 1},
    Chongyu Liu\textsuperscript{\rm 1},
    Lianwen Jin\textsuperscript{\rm 1\rm 2}\thanks{Corresponding author}
}
\affiliations{
    \textsuperscript{\rm 1}South China University of Technology\\
    \textsuperscript{\rm 2}INTSIG-SCUT Joint Lab on Document Analysis and Recognition\\
    \{eezhyang, pengdezhi000, scut.yxshi, yuyizhang.scut, liuchongyu1996\}@gmail.com, 
    eelwjin@scut.edu.cn

}

\usepackage{bibentry}
\begin{document}

\maketitle

\begin{abstract}
    Historical documents encompass a wealth of cultural treasures but suffer from severe damages including character missing, paper damage, and ink erosion over time.
    However, existing document processing methods primarily focus on binarization, enhancement, etc., neglecting the repair of these damages.
    To this end, we present a new task, termed Historical Document Repair (\textbf{HDR}), which aims to predict the original appearance of damaged historical documents.
    To fill the gap in this field, we propose a large-scale dataset \textbf{HDR28K} and a diffusion-based network \textbf{DiffHDR} for historical document repair.
    Specifically, HDR28K contains 28,552 damaged-repaired image pairs with character-level annotations and multi-style degradations.
    Moreover, DiffHDR augments the vanilla diffusion framework with semantic and spatial information and a meticulously designed character perceptual loss for contextual and visual coherence.
    Experimental results demonstrate that the proposed DiffHDR trained using HDR28K significantly surpasses existing approaches and exhibits remarkable performance in handling real damaged documents.
    Notably, DiffHDR can also be extended to document editing and text block generation, showcasing its high flexibility and generalization capacity.
    We believe this study could pioneer a new direction of document processing and contribute to the inheritance of invaluable cultures and civilizations.
    The dataset and code is available at https://github.com/yeungchenwa/HDR.
\end{abstract}


\section{Introduction}

\begin{figure}[t]
    \centering
    \includegraphics[width=0.8\columnwidth]{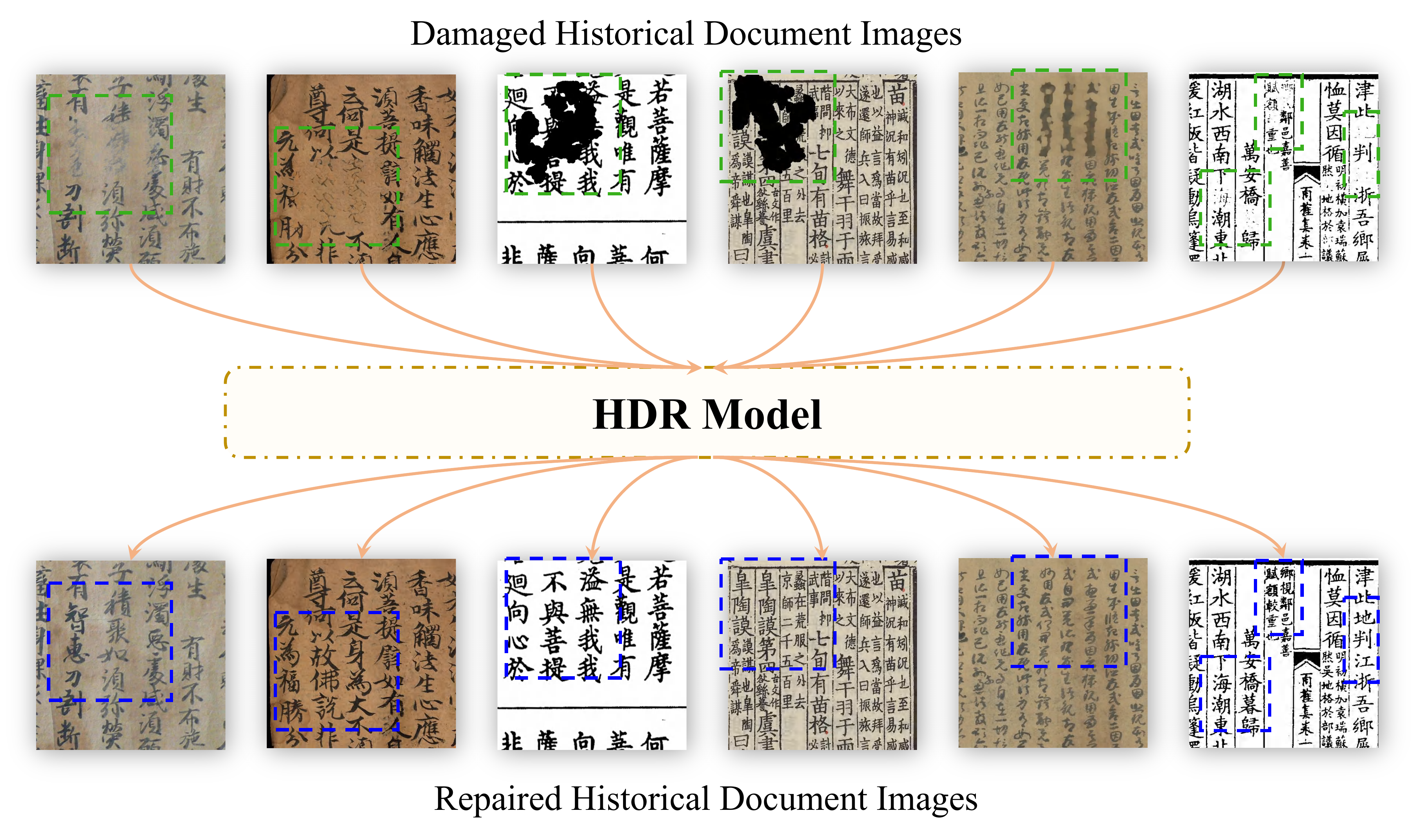}
    \caption{Definition of Historical Document Repair (HDR) task. The green boxes represent the damaged regions and the blue boxes denote the repaired regions.}
    \label{fig:task_definition}
\end{figure}

% 先说古籍的重要性
Historical documents play a pivotal role in the transmission of cultural heritage.
However, during prolonged preservation, they are susceptible to oxidization, insect damage, water erosion, \textit{etc}., leading to character missing, paper damage, and ink erosion, as shown in Figure \ref{fig:task_definition}.
Nevertheless, the manual repair of damaged characters and corrupted backgrounds is a complex and time-consuming endeavor.

% 再说古籍修复的重要性 
Recently, generic document image processing primarily concentrates on the low-level vision tasks, such as rectification \cite{li2023foreground,jiang2022revisiting}, binarization \cite{yang2023docdiff,yang2023novel}, enhancement \cite{hertlein2023template,wang2022udoc,xue2022fourier}, and deshadowing \cite{li2023high,lin2020bedsr}.
However, they fall short of understanding the semantics and stylistic elements within document images, thereby hindering their capability to repair the damaged documents.
Moreover, existing historical document processing methods also address tasks such as text restoration \cite{assael2022restoring} and individual character restoration \cite{nguyen2019character,amin2023reconstruction}; however, these unimodal methods are unsuitable for document repair because predicting the original appearance of damaged documents is a highly challenging multimodal task, requiring the understanding of the context and the pixel-level restoration.
Though the recent work \cite{zhu2024reproducing} conducts inscription restoration task, it specifically targets inscriptions with simple backgrounds, exclusively characterized by white text on a black background.
Additionally, the font generation task \cite{kong2022look,wang2023cf,yang2023fontdiffuser} exhibits a higher resemblance, involving character generation under the conditions of content and style. 
Nevertheless, this task is only employed for individual characters and it is not feasible to reconstruct the documents.

% 任务的定义
Therefore, to fill the gap in this field, we introduce a new task, termed \textit{\underline{H}istorical \underline{D}ocument \underline{R}epair} (\textbf{HDR}), which involves predicting the original appearance of damaged historical document images. 
As shown in Figure \ref{fig:task_definition}, the damaged historical document images are fed into the HDR model to repair the damaged regions.
The output images of HDR model, termed repaired images, should not only capture precise character content and style but also harmonize with the surrounding background within the repaired region.

\begin{figure}
    \centering
    \includegraphics[width=0.9\columnwidth]{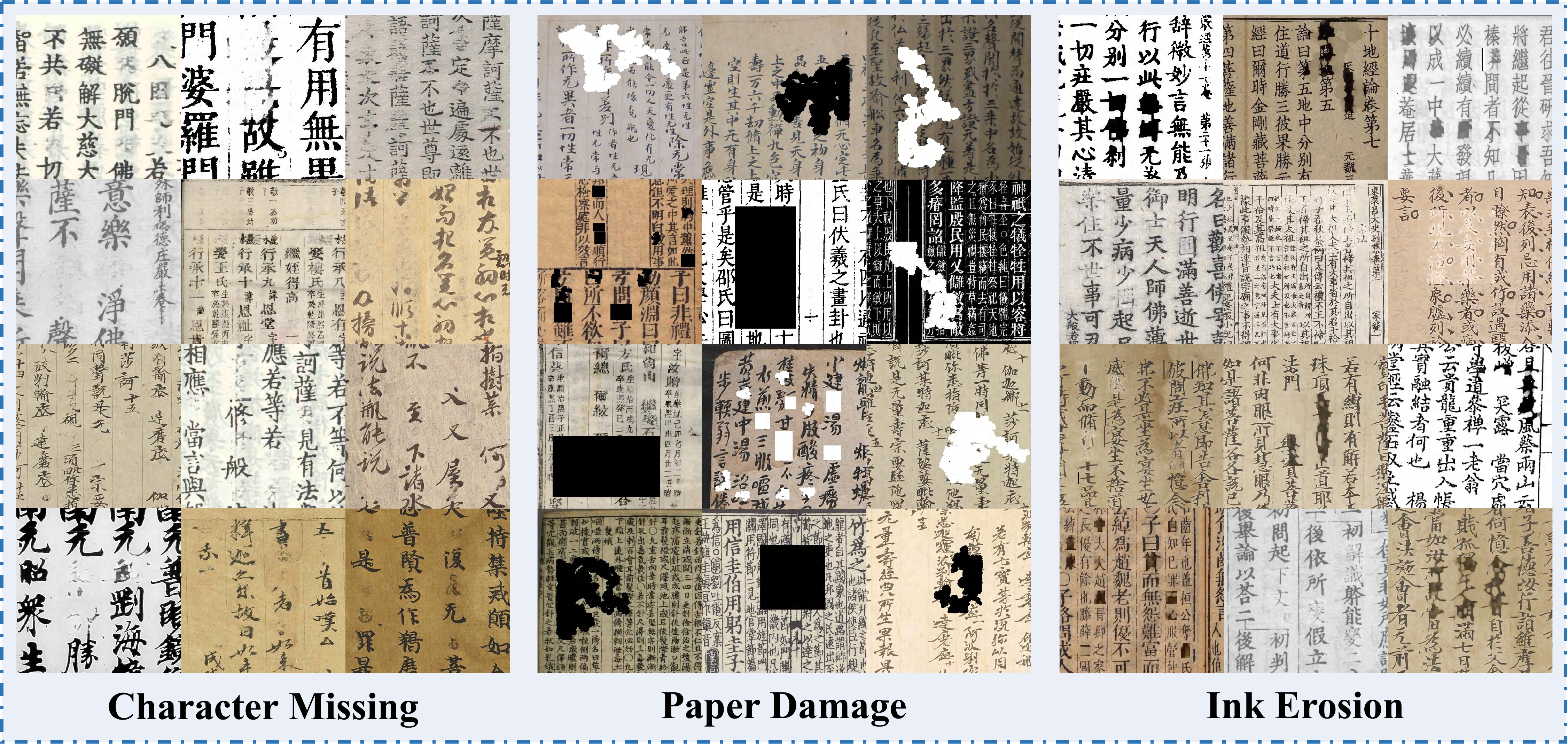}
    \caption{Damaged Samples in HDR28K.}
    \label{fig:degraded_samples}
\end{figure}

% 提出我们的数据集
As there is no dataset available for historical document repair, we contribute a large-scale dataset, named \textbf{HDR28K}, which comprises a total of 28,552 damaged-repaired image pairs with character-level annotations and multi-style degradation.
As shown in Figure \ref{fig:degraded_samples}, the undamaged images are corrupted by three meticulously designed degradations to simulate character missing, paper damage, and ink erosion, which is intended to faithfully replicate the visual effects of damages observed in historical documents.

% 提出我们的方法
Additionally, to facilitate the development of HDR task, we propose \textbf{DiffHDR}, a \underline{Diff}usion-based \underline{H}istorical \underline{D}ocument \underline{R}epair network, which frames the HDR task as a series of diffusion steps that progressively transform the damaged regions to match the target character content and character style with an accurate background. 
In our method, we first crop the damaged region from the historical document to obtain a fixed-size damaged patch image, then DiffHDR leverages the damaged images, along with semantic and spatial information, as conditions for appearance reconstruction. 
To further improve the content preservation of repaired characters, we introduce a character perceptual loss to penalize the misalignment of character features. 

% 实验结果 
Extensive experiments demonstrate that the models trained using HDR28K can reconstruct the original appearance of damaged historical document images and achieve state-of-the-art performance.
Moreover, we have gathered a collection of real damaged samples from the Internet and applied our method for repair, which shows that DiffHDR, trained on synthetic data, is proficient in real scenarios, highlighting its significant potential for the preservation of cultural heritage.
Furthermore, DiffHDR can be extended to document editing and text block font generation, exhibiting the flexibility and generalization of our proposed method.

We summarize our main contributions as follows:
\begin{itemize}
\item[$\bullet$] 
We introduce a Historical Document Repair (HDR) task, which endeavors to predict the original appearance of damaged historical document images.

\item[$\bullet$]
We build a large-scale historical document repair dataset, termed HDR28K, which includes 28,552 damaged-repaired image pairs with character-level annotations and multi-style degradation.

\item[$\bullet$]
We propose a \underline{Diff}usion-based \underline{H}istorical \underline{D}ocument \underline{R}epair method (DiffHDR), which augments the DDPM framework with semantic and spatial information and incorporates a meticulously designed character perceptual loss to enhance the contextual and visual coherence.

\item[$\bullet$]
DiffHDR trained on HDR28K outperforms other methods and is capable of repairing real damaged historical documents. Moreover, our method can be extended to document editing and text block font generation.
\end{itemize} 

\section{Related Work}

\subsection{Image Restoration}
Generic image restoration methods \cite{li2023efficient,cui2023focal,chen2022simple,wang2022uformer,zamir2022restormer} are primarily focused on image deraining, defogging, deblurring, or denoising. 
For example, Restormer \cite{zamir2022restormer} proposes an efficient transformer model to capture long-range pixel interactions and is applicable to large images. 
In document image restoration, they mainly address the tasks of rectification \cite{li2023foreground,jiang2022revisiting,das2019dewarpnet}, binarization \cite{yang2023docdiff,yang2023novel}, enhancement \cite{hertlein2023template,wang2022udoc}, and deshadowing \cite{li2023high,lin2020bedsr}. 
Although the above methods have achieved remarkable performance, they cannot comprehend the semantics and stylistic elements present in document images, thereby hindering the repair of damaged documents.

\subsection{Historcial Document Image Processing}
Some approaches have been proposed for historical document image processing. 
Ithaca \cite{assael2022restoring} utilizes the transformer block to conduct the sequence modeling for textual restoration, geographical attribution, and chronological attribution. 
Some methods \cite{nguyen2019character, amin2023reconstruction} focus on individual character restoration. 
\cite{amin2023reconstruction} focuses on the isolated Greek characters and applies an auto-encoder to reconstruct the missing parts of characters. 
Moreover, to alleviate the unreadability of damaged historical documents, \cite{ech2022frank} propose a multi-task learning module to conduct the binarization task. 
Furthermore, some methods \cite{hedjam2013historical,raha2019restoration,wadhwani2021text} are proposed to conduct a historical document image enhancement. 
% For example, \cite{hedjam2013historical} introduces an approach to provide the historian with an acceptable visualization of the degraded historical document images. 
Nevertheless, these unimodal methods prove inadequate for document repair, as historical document repair is a highly challenging multimodal task, which demands an understanding of both the context and pixel-level restoration. 
The recent work \cite{zhu2024reproducing} introduces an inscription restoration dataset; however, it lacks diversity in styles.
Thus, to fill the gap in this field, we contribute a large-scale historical document repair dataset with diverse complex backgrounds, termed HDR28K, and propose a diffusion-based model.

\section{Historical Document Repair}
The objective of historical document repair (HDR) is to accurately predict the original appearance of damaged historical document images. 
Specifically, as illustrated in Figure \ref{fig:task_definition}, when presented with a damaged historical document image $\boldsymbol{x}_d$, the HDR model focuses on reconstructing the original state, obtaining the repaired image $\boldsymbol{x}_{r}$.
This reconstruction process requires the HDR model to repair both the damaged characters and the corrupted background. 
The repaired output should not only precisely capture the content and style of characters but also seamlessly integrate with the surrounding background within the repaired region.
Thus, the repaired result $\boldsymbol{x}_r$ can be formulated as follows:
\begin{align}
    \boldsymbol{x}_r = \mathcal{F}_{HDR}(\boldsymbol{x}_d). 
\end{align}
$\mathcal{F}_{HDR}$ denotes the historcial document repair model. In our work, we leverage the priors of semantic and spatial cues (provided by our HDR28K dataset) to support the repair of damaged historical documents. Thus, the repair of our method is as follows:
\begin{align}
    \boldsymbol{x}_r = \mathcal{F}(\boldsymbol{x}_d, \boldsymbol{x}_c, \boldsymbol{x}_m),
\end{align}
where $\boldsymbol{x}_c$ represents the content image (senmantic prior) and $\boldsymbol{x}_m$ denotes the mask image (spatial prior).

\section{HDR28K Dataset}
As there is no dataset specifically designed for historical document repair, we construct HDR28K, which consists of 28,552 damaged-repaired image pairs with OCR annotations. In this section, we provide the construction details and analysis of the proposed dataset.

\begin{figure}
    \centering
    \includegraphics[width=\linewidth]{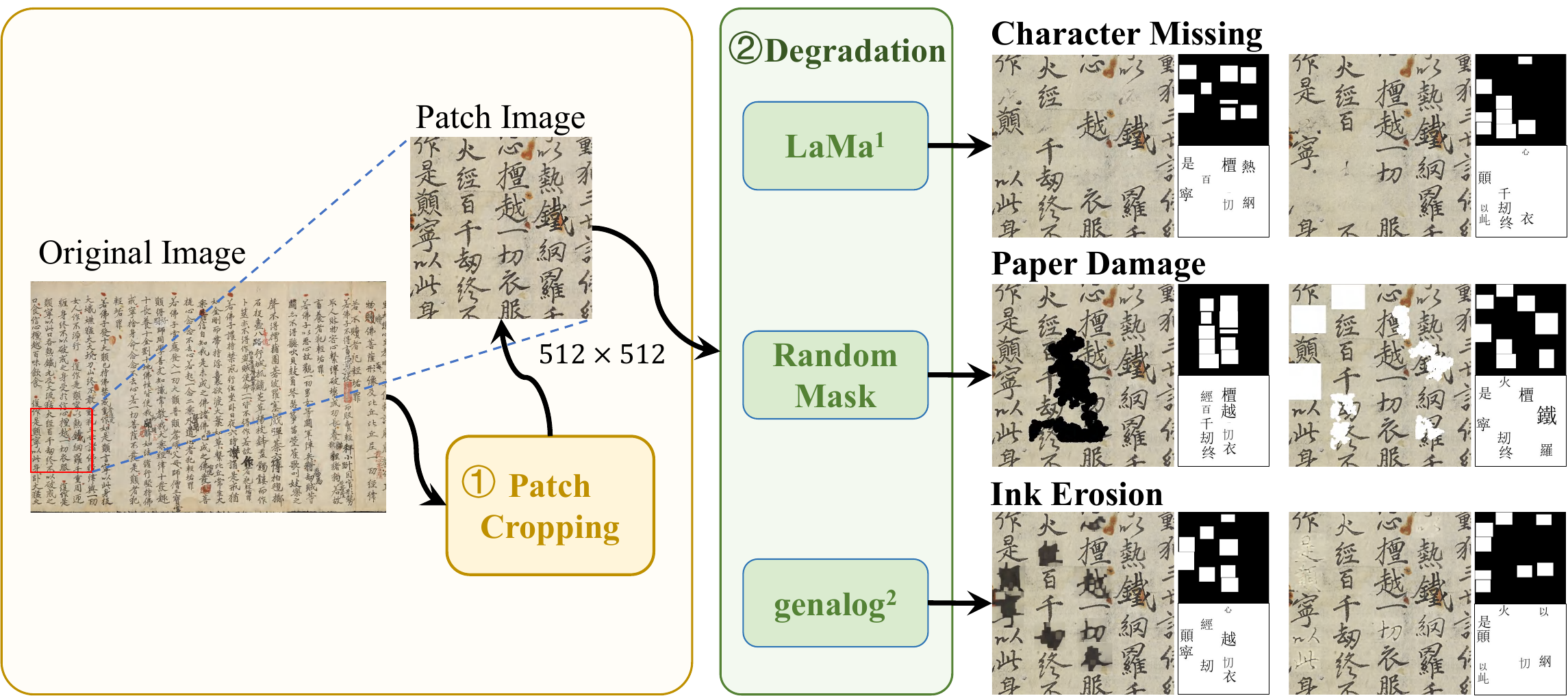}
    \caption{Construction pipeline of the HDR28K dataset.}
    \label{fig:degradation_pipeline}
\end{figure}

\begin{figure}[h]
    \centering
    \subfloat[Sample Distribution]{
        \includegraphics[width=0.50\columnwidth]{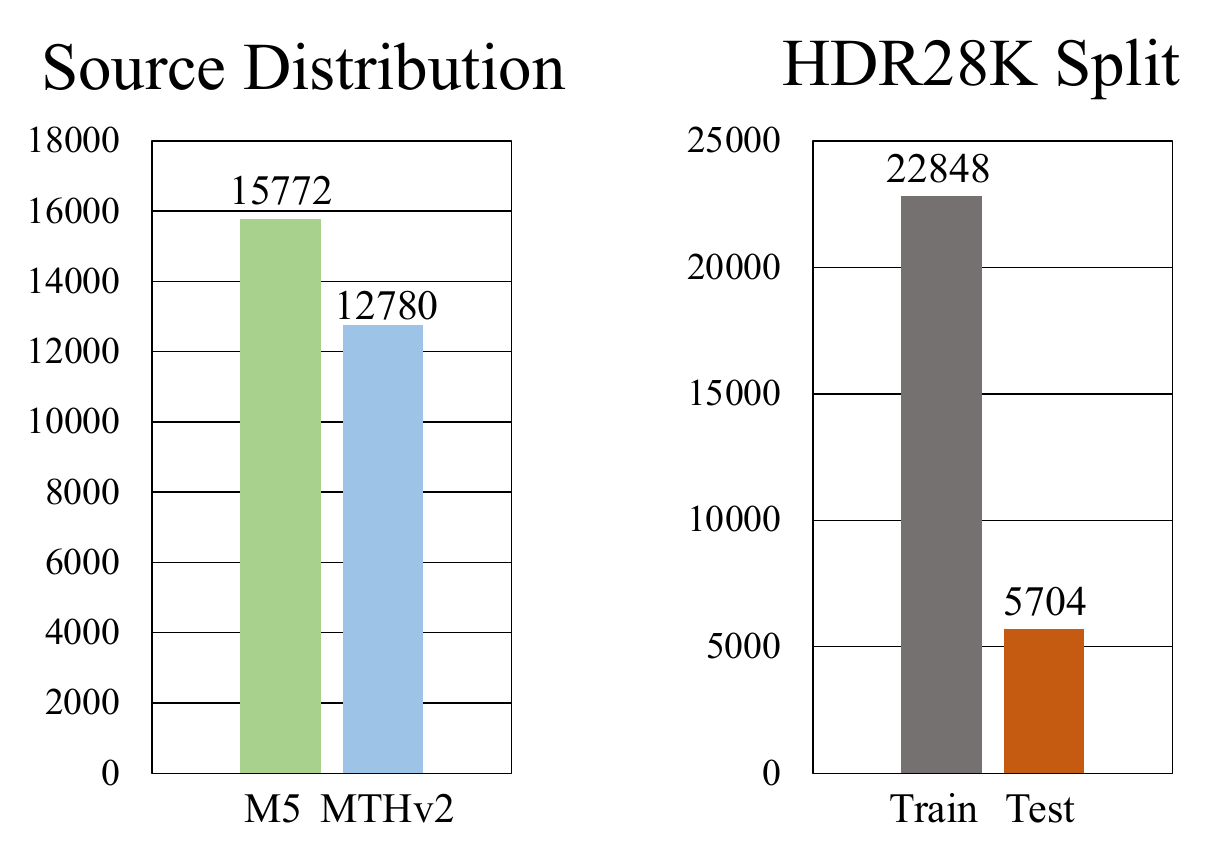}}
    \hfill
    \subfloat[Degradation Distribution]{
        \includegraphics[width=0.45\columnwidth]{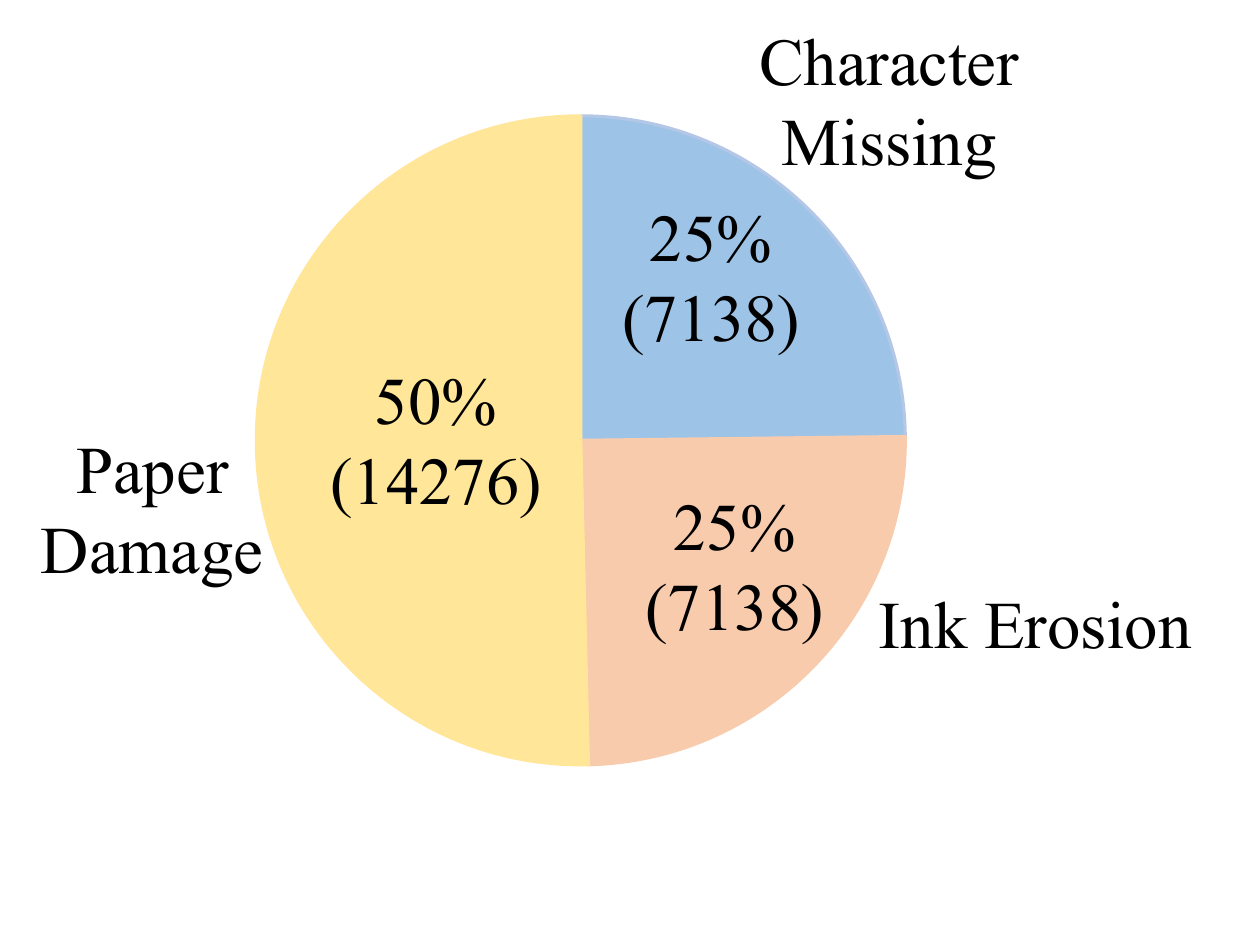}}
    \caption{Statistics of the HDR28K dataset.}
    \label{fig:data_analysis}
\end{figure}

\subsection{Data Construction}

\begin{figure*}[t]
    \centering
    \includegraphics[width=0.96\textwidth]{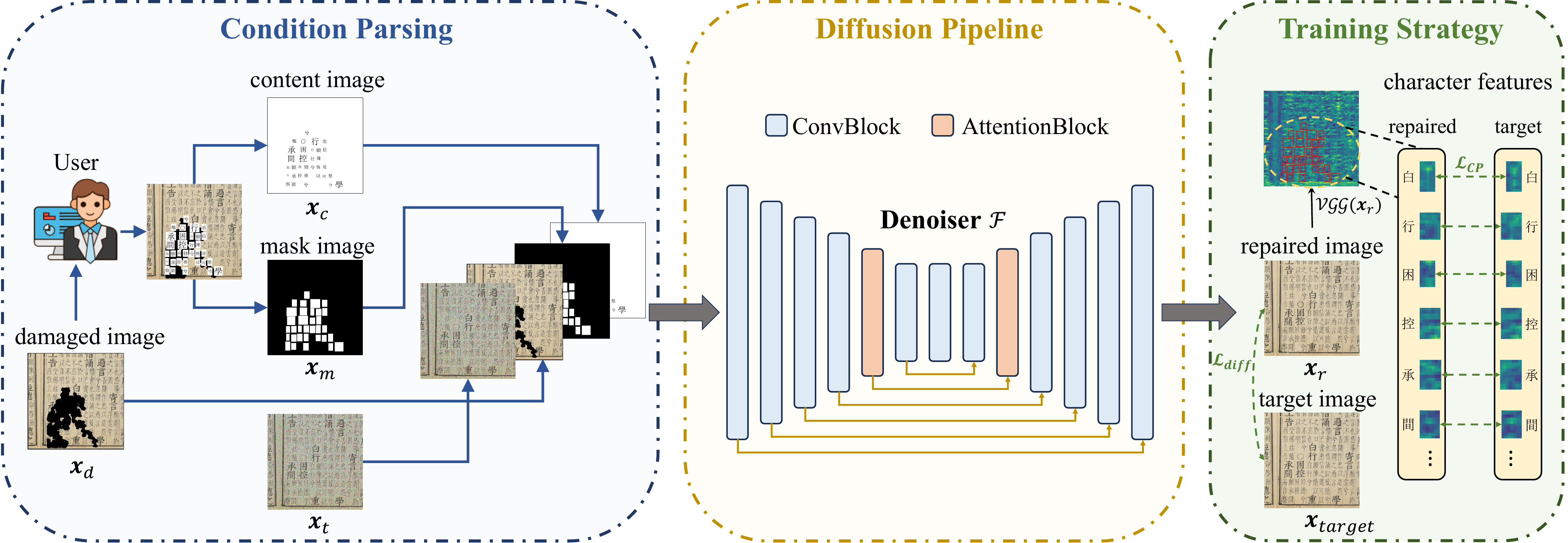}
    \caption{Overview of our proposed method. DiffHDR comprises a condition parsing and a diffusion pipeline. 
    In the condition parsing, the user provides the content and location of damaged characters, obtaining the content image $\boldsymbol{x}_{c}$ and mask image $\boldsymbol{x}_{m}$. 
    In the diffusion pipeline, our denoiser $\mathcal{F}$, a UNet-based network, outputs the repaired image $\boldsymbol{x}_{r}$ conditioned on noised image $\boldsymbol{x}_{t}$, damaged image $\boldsymbol{x}_{d}$, mask image $\boldsymbol{x}_{m}$ and content image $\boldsymbol{x}_{c}$.
    During training, in addition to using diffusion loss $\mathcal{L}_{diff}$, we introduce a character perceptual loss $\mathcal{L}_{CP}$ to enhance the content preservation of repaired characters.}
    \label{fig:framework}
\end{figure*}

% To construct HDR28K dataset, we collect samples from MTHv2 \cite{ma2020joint} and M5HisDoc \cite{shi2023m5hisdoc} (two Chinese historical document image datasets), and implement meticulously designed degradations on undamaged images. 
To accurately simulate real damaged scenarios in historical documents, it is necessary to degrade the characters, which requires character-level bounding boxes and content annotations. 
Therefore, we constructed the HDR28K dataset by building upon MTHv2 \cite{ma2020joint} and M5HisDoc \cite{shi2023m5hisdoc} and implementing meticulously designed degradations.
Specifically, as illustrated in Figure \ref{fig:degradation_pipeline}, for efficiency in memory and computation, we first crop $512\times 512$ patch images from high-resolution original images using sliding windows. 
During cropping, our automated schemes focus exclusively on text regions, and we manually filter out the images that are low resolution or lack text intensity.
Subsequently, we apply three degradations on patch images, which replicate the real scenarios of character missing, paper damage, and ink erosion in damaged situations.

The details of the three degradations are:
\textbf{(1) Character Missing}: 
We randomly generate masks and employ LAMA \cite{suvorov2022resolution} to erase the content in mask regions. 
The generated masks consist of character-level and block-level types. 
Because MTHv2 and M5HisDoc provide the location annotations for corresponding individual character regions, we randomly select some of these character regions as the character-level masks.
In the generation of block-level masks, we randomly sample a rectangular region from the patch image as the erasing mask. 
\textbf{(2) Paper Damage}: 
Due to insect infestation, oxidization, contamination, etc., the papers in historical documents suffer from severe damage. 
To replicate this scenario, we randomly mask some regions in the patch image using black or white pixels. 
Similar to the character missing, the masked regions include character-level and block-level types, but they take the form of either a rectangular or an irregular shape.
\textbf{(3) Ink Erosion}: 
We utilize genalog\footnote{https://github.com/microsoft/genalog} to simulate scenarios involving water erosion and character fading.
We first randomly sample rectangular regions from patch images similar to the mask generation in character missing.
Then we apply diverse degradation modes and convolution kernels in genalog to induce degradation to the sampled regions. 
The examples of the above three degradations are listed on the right of Figure \ref{fig:degradation_pipeline}.

\subsection{Data Analysis}

We randomly select 536 original images from the testing set of MTHv2 and 891 original images from the testing set of M5HisDoc to construct our HDR28K testing set. 
The HDR28K training set is sourced from the remaining samples in the two datasets. 
Note that the patch images from the same historical documents are not assigned to both training and testing sets.
As shown in Figure \ref{fig:data_analysis}(a), after the cropping in the construction pipeline, the training set in HDR28K comprises 22,848 patch images, while the testing set consists of 5,704 patch images. 
Moreover, 12,780 patch images originate from MTHv2 \cite{ma2020joint} while 15,772 patch images are sourced from M5HisDoc \cite{shi2023m5hisdoc}. 
As depicted in Figure \ref{fig:data_analysis}(b), the degradation of paper damage accounts for 50\% of the HDR dataset, while the other two degradations account for 25\%, respectively.
Finally, we present some samples in Figure \ref{fig:degraded_samples}, which demonstrates that our dataset can realistically replicate the damage observed in historical document images. 

\section{DiffHDR: Diffusion-based HDR Network}

\subsection{Framework}
As depicted in Figure \ref{fig:framework}, the framework of DiffHDR consists of a condition parsing and a diffusion pipeline. During condition parsing, the user provides the content and location of the damaged characters and we parse them out to obtain the content image $\boldsymbol{x}_c$ and the mask image $\boldsymbol{x}_m$. Subsequently, our diffusion pipeline gathers the damaged image $\boldsymbol{x}_{d}$ with the user's guidance to predict the original appearance $\boldsymbol{x}_r$ of the damaged image $\boldsymbol{x}_d$.

Specifically, we randomly sample a time step $t\sim Uniform(0, T_{max})$ and a Gaussian noise $\boldsymbol{\epsilon}_{t}$ to corrupt the damaged image $x_{0}$, yielding the noised image $\boldsymbol{x}_{t}$ following \cite{ho2020denoising}:
\begin{align}
    \boldsymbol{x}_{t} = \sqrt{\bar{\alpha}_{t}}\boldsymbol{x}_{0}+\sqrt{1-\bar{\alpha}_{t}}\boldsymbol{\epsilon},  
\end{align}
where $\alpha_{t}=1-\beta_{t}$, $\bar{\alpha}_{t} = \prod_{i=0}^{t} (1 - \beta_{i})$, $\beta_{i} \sim (0, 1)$.
Then we concatenate $\boldsymbol{x}_t \sim \mathbb{R}^{3 \times H \times W}$, $\boldsymbol{x}_d \sim \mathbb{R}^{3 \times H \times W}$, $\boldsymbol{x}_c \sim \mathbb{R}^{1 \times H \times W}$ and $\boldsymbol{x}_m \sim \mathbb{R}^{1 \times H \times W}$ in channel dimension as a 8-channel input to the following denoiser $\mathcal{F}$. Our denoiser $\mathcal{F}$ is a UNet-based network, which directly predicts the repaired result $\boldsymbol{x}_r$ rather than the added noise $\boldsymbol{\epsilon}_{t}$. 
In this manner, our method is capable of performing pixel-level repairs, significantly reducing both labor and time costs while advancing the field of digital humanities.

\subsection{Training Objective}
\noindent \textbf{Diffusion Loss}
Because our proposed method directly predicts the original appearance $\boldsymbol{x}_{r}$ of the damaged images rather than the added noise $\boldsymbol{\epsilon}_{t}$, we optimize DiffHDR with the diffusion loss as follows:
\begin{align}
    \mathcal{L}_{diff} = \left \| \boldsymbol{x}_{target} - \mathcal{F}(\boldsymbol{x}_t; \boldsymbol{x}_d, \boldsymbol{x}_c, \boldsymbol{x}_m) \right \|^{2},
\end{align}
where $\boldsymbol{x}_{target}$ denotes the target image. 

\noindent \textbf{Character Perceptual Loss} 
To further improve the content preservation of repaired characters, we introduce a Character Perceptual Loss (CPLoss) to provide guidance to our model. 
Specifically, as shown in right of Figure \ref{fig:framework}, we first utilize the pretrained VGG \cite{simonyan2014very} to extract the feature $\mathcal{VGG}(\boldsymbol{x}_r)$ from the repaired image $\boldsymbol{x}_r$. 
Then we penalize the misalignment of feature between $\mathcal{VGG}(\boldsymbol{x}_r)$ and the target feature $\mathcal{VGG}(\boldsymbol{x}_{target})$ within the repaired regions. The CPLoss is formulated as follows:
\begin{align}
    \mathcal{L}_{CP} = \sum_{i=1}^{L} \omega_{i}(\left \|\mathcal{VGG}_{i}(\boldsymbol{x}_{r})-\mathcal{VGG}_{i}(\boldsymbol{x}_{target}) \right \|) \boldsymbol{x}_m,
\end{align}
where $\mathcal{VGG}_{i}$ represents the $i$-th VGG layer feature. 
$\omega_{i}$ denotes the layer weight. 
To effectively capture both global and local representations of repaired characters, we utilize multi-scale features to penalize the misalignment.
$\boldsymbol{x}_{m}$ enables the concentration of DiffHDR solely on damaged characters.
CPLoss not only ensures the preservation of content and style for characters within the repaired regions but maintains the compatibility of the repaired background as well.

\subsection{Attribute-Sensitive Repair Strategy}
In HDR task, our denoiser $\mathcal{F}$ has three attributes: the damaged image $\boldsymbol{x}_{d}$, content image $\boldsymbol{x}_{c}$ and mask image $\boldsymbol{x}_{m}$. 
Inspired by InstructPix2Pix \cite{brooks2023instructpix2pix}, it is beneficial to utilize the classifier-free guidance \cite{ho2022classifier} in relation to the conditional inputs. 
Therefore, during training, we randomly set only $\boldsymbol{x}_{d}=\emptyset$, both $\boldsymbol{x}_{c}=\emptyset$ and $\boldsymbol{x}_{m}=\emptyset$, and all $\boldsymbol{x}_{d}=\emptyset$, $\boldsymbol{x}_{c}=\emptyset$ and $\boldsymbol{x}_{m}=\emptyset$ with an 8\% probability, respectively (where $\emptyset$ indicates setting the unconditional inputs to pixel values of 255 or 0). 
This strategy enables our method more sensitive to the three attributes. 

During sampling, we introduce the guidance scales $s_{d}$ and $s_{c,m}$, which can be viewed as the sensitivity of the repaired results with the damaged image $\boldsymbol{x}_{d}$ and the content and location cues $\boldsymbol{x}_{c}$, $\boldsymbol{x}_{m}$, respectively. Thus, the repair strategy can be formulated as:
\begin{align}
    \tilde{\mathcal{F}}(\boldsymbol{x}_t; &\boldsymbol{x}_d, \boldsymbol{x}_c, \boldsymbol{x}_m) = \mathcal{F}(\boldsymbol{x}_t; \emptyset, \emptyset, \emptyset) \notag \\
    &+ s_{d}(\mathcal{F}(\boldsymbol{x}_t; \boldsymbol{x}_d, \emptyset, \emptyset)-\mathcal{F}(\boldsymbol{x}_t; \emptyset, \emptyset, \emptyset)) \notag \\
    &+ s_{c.m}(\mathcal{F}(\boldsymbol{x}_t; \boldsymbol{x}_d, \boldsymbol{x}_c, \boldsymbol{x}_m)-\mathcal{F}(\boldsymbol{x}_t; \boldsymbol{x}_d, \emptyset, \emptyset)).
\end{align}

\begin{table}[]
    \centering
    \resizebox{\columnwidth}{!}{%
    \begin{tabular}{@{}l|l|ccc@{}}
    \toprule
    \multicolumn{1}{c|}{Model} & Venue    & FID$\downarrow$            & LPIPS$\downarrow$           & Rec-ACC(\%)$\uparrow$         \\ \midrule
    UNet \cite{ronneberger2015u}                      & MICCAI'15        & 1.5504          & 0.0638          & 55.5547         \\
    Pix2Pix-ResNet \cite{isola2017image}             & CVPR'17 & 7.9586          & 0.0821          & 38.7583         \\
    Pix2Pix-UNet \cite{isola2017image}               & CVPR'17 & 12.4989         & 0.0816          & 39.7935         \\
    CycleGAN-ResNet \cite{zhu2017unpaired}           & ICCV'17 & 4.6521          & 0.0935          & 33.2306         \\
    CycleGAN-UNet \cite{zhu2017unpaired}             & ICCV'17 & 12.4847         & 0.1192          & 27.5988         \\
    Uformer-Tiny \cite{wang2022uformer}              & CVPR'22 & 1.4743          & 0.0626          & 57.7111         \\
    Uformer-Small \cite{wang2022uformer}             & CVPR'22 & 1.1858          & 0.0547          & 67.5760         \\
    Uformer-Big \cite{wang2022uformer}               & CVPR'22 & 1.026           & 0.0510          & \underline{69.9545}   \\
    Restormer \cite{zamir2022restormer}                 & CVPR'22 & 1.1632          & 0.0584          & 60.4338         \\
    NAFNet \cite{chen2022simple}                    & ECCV'22 & \underline{0.8588}    & \underline{0.0435}    & 69.5821         \\
    GRL \cite{li2023efficient}                       & CVPR'23 & 3.5562          & 0.0819          & 41.0406         \\
    FocalNet \cite{cui2023focal}                  & ICCV'23 & 14.7901         & 0.1034          & 44.0407         \\
    UPOCR \cite{peng2023upocr}                 & ICML'24 & 6.5690           & 0.0903          & 40.1415         \\
    Ours                       & AAAI'25        & \textbf{0.7499} & \textbf{0.0384} & \textbf{81.9180} \\ \bottomrule
    \end{tabular}%
    }
    \caption{Quantitative comparison. The bold indicates the state-of-the-art and the underline indicates the second best.}
    \label{tab:quantitative_result}
\end{table}

\section{Experiment}
\subsection{Evaluation Metrics}
We utilize FID \cite{heusel2017gans}, LPIPS \cite{zhang2018unreasonable}, and the accuracy of character recognizer (Rec-ACC) for quantitative comparison. 
FID measures the distribution distance between the output and the target domain.
LPIPS is closer to human visual perception. 
We utilize a trained character recognizer to evaluate the accuracy of the individual characters within the repaired regions in $\boldsymbol{x}_{r}$ and the result is called Rec-ACC.
We employ VGG19 \cite{simonyan2014very} as the character recognizer and it is trained using all individual character data in MTHv2 and M5HisDoc.
Because HDR task focuses on the repaired region, we replace the non-damaged region of $\boldsymbol{x}_{r}$ by the target $\boldsymbol{x}_{target}$ before the evaluation. In short, FID and LPIPS are the image-level metrics, while Rec-ACC is the instance-level metric.
Additionally, we do not evaluate two commonly used metrics, PSNR and SSIM, as they are unsuitable for HDR task \cite{zhang2018unreasonable}. 
The evidence is provided in Section 6.3.

\begin{figure}
    \centering
    \includegraphics[width=0.8\linewidth]{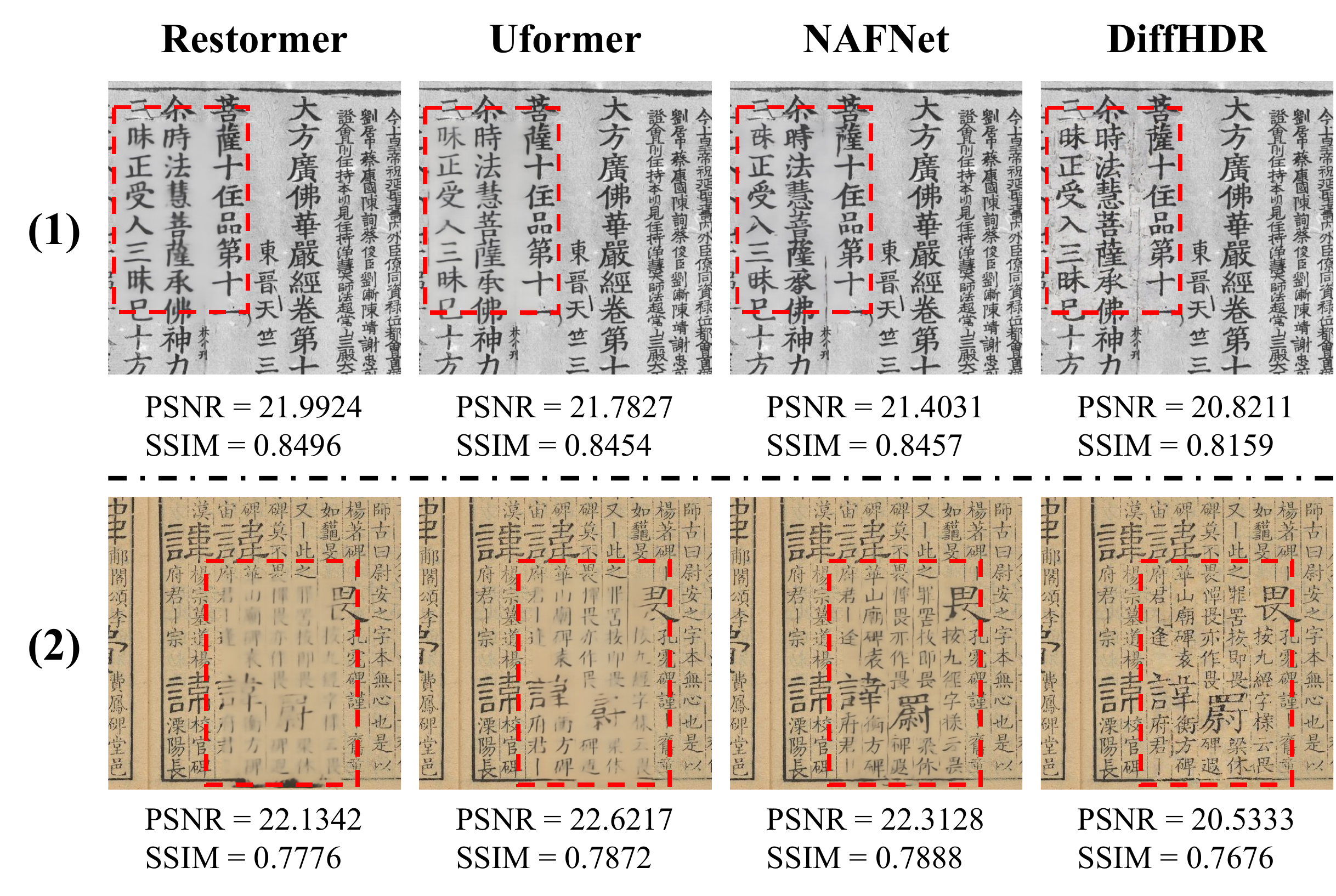}
    \caption{Unsuitableness of PSNR and SSIM.}
    \label{fig:psnr_ssim}
\end{figure}

\subsection{Implementation Details}
We adopt an AdamW optimizer to train DiffHDR with $\beta_{1}=0.95$ and $\beta_{2}=0.999$. 
The image size is $512\times 512$. 
During classifier-free training, we set the conditional dropout probability as $8$\% and we train the model with a batch size of $32$ and a total epoch of $165$. 
The learning rate is set as $1\times 10^{-4}$ with the linear schedule. 
The training is conducted on $8$ NVIDIA A6000 GPUs. 
We set the guidance scales $s_{d}=1.2$ and $s_{c,m}=1.5$ and adopt the DPM-Solver++ \cite{lu2022dpm} as our sampler with the inference step of 20.

\subsection{Comparison with Existing Methods}
We compare our method with 9 methods, including GAN-based methods (Pix2Pix and CycleGAN), CNN-based methods (UNet, NAFNet, and FocalNet), and Transformer-based methods (Uformer, Restormer, GRL, and UPOCR). 
Since these existing methods are not originally designed for the HDR task, they are adapted to use the concatenation of damaged image $\boldsymbol{x}_{d}$, content image $\boldsymbol{x}_{c}$, and mask image $\boldsymbol{x}_{m}$ as a 5-channel input and generate a 3-channel repaired image $\boldsymbol{x}_r$ as output.
For a fair comparison, all these methods are trained on $8$ NVIDIA A6000 GPUs with the same batch size and epochs as DiffHDR. 
Moreover, we adopt ResNet or UNet as the generator of Pix2Pix and CycleGAN, as shown in the 2\textit{nd} to 5\textit{th} rows of Table~\ref{tab:quantitative_result}.
Note that we do not conduct the experiments on the CIRI dataset \cite{zhu2024reproducing}, nor do we use its method for comparison, as they have not yet been made open-source.

\subsubsection{Quantitative Comparison}

The quantitative results are shown in Table \ref{tab:quantitative_result}. 
DiffHDR achieves state-of-the-art performance, surpassing other methods by a substantial margin in FID, LPIPS and Rec-ACC. 
Our method achieves a 12.7\% lower FID and an 11.7\% lower LPIPS compared to the second-best NAFNet.
Notably, DiffHDR outperforms the second-best Uformer-Big by 11.9635\% in Rec-ACC, highlighting its advantage in character correctness. 
We find that PSNR and SSIM are unsuitable for HDR task. 
Supporting evidence is presented in Figure \ref{fig:psnr_ssim}, where blurring causes higher values of PSNR and SSIM \cite{zhang2018unreasonable}.

\begin{figure}
    \centering
    \subfloat[Document editing.]{
        \includegraphics[width=0.47\columnwidth]{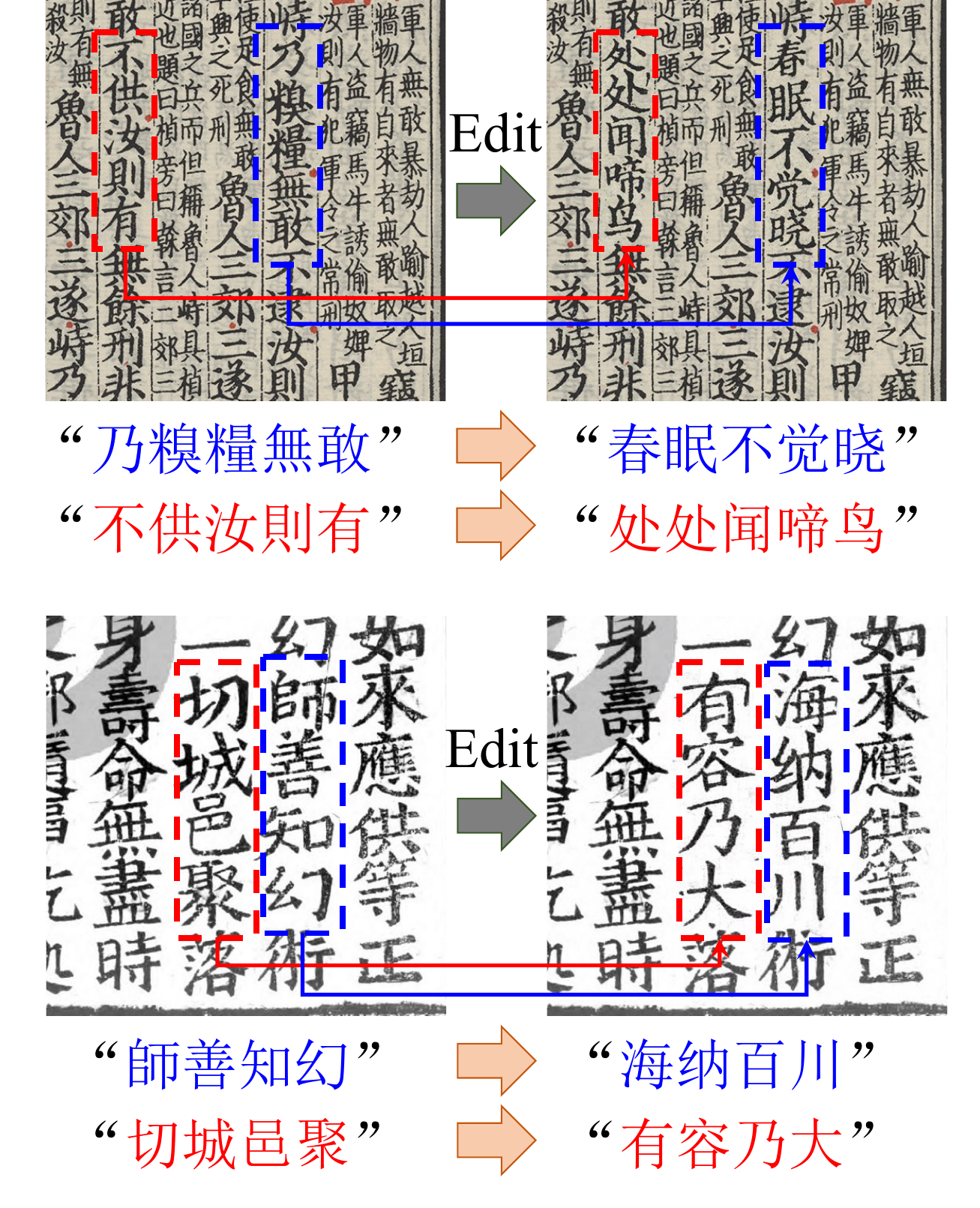}}
    \hfill
    \subfloat[Text Block Font Generation.]{
        \includegraphics[width=0.48\columnwidth]{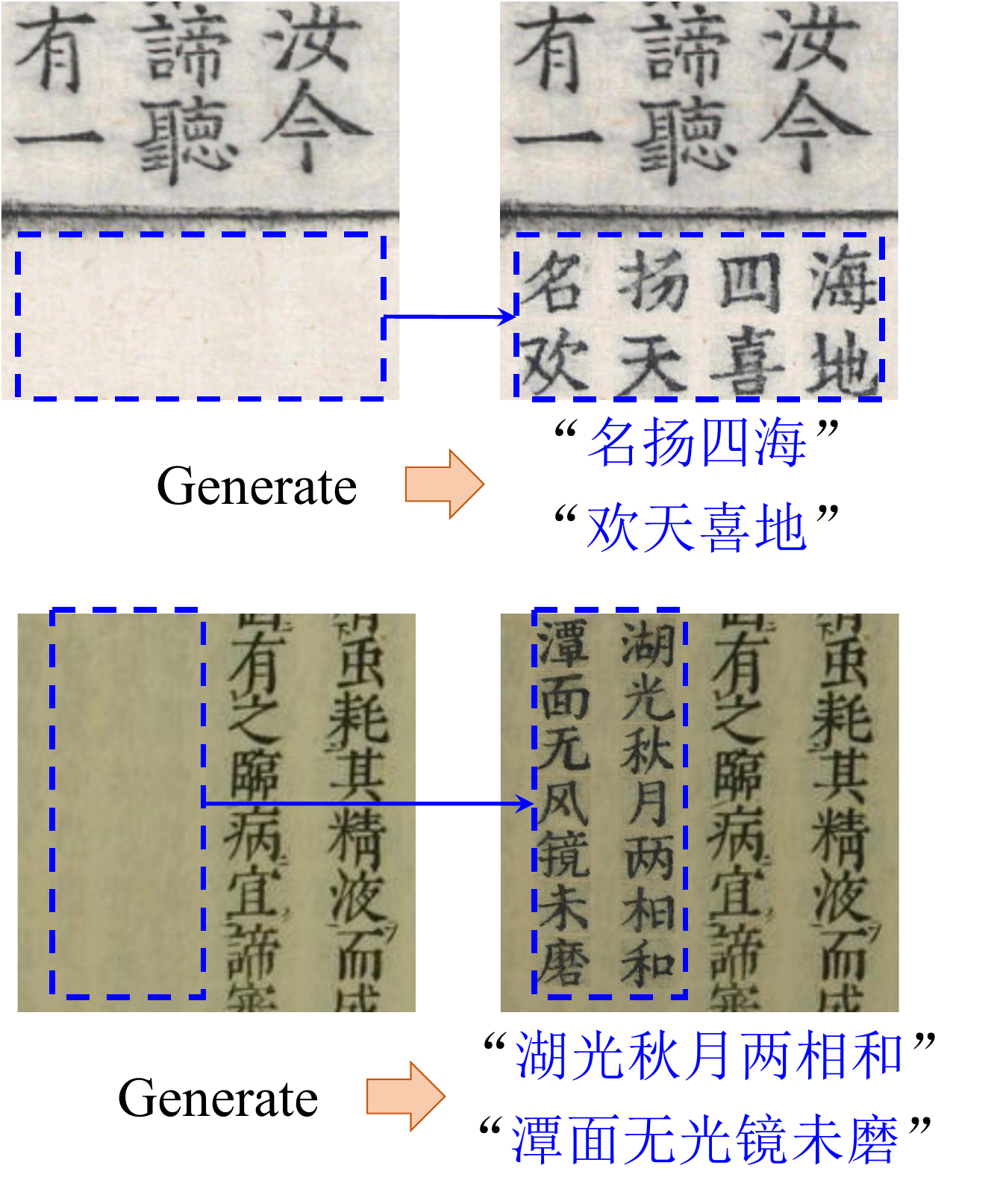}}
    \caption{Document editing and text block font generation.}
    \label{fig:other_application}
\end{figure}

\begin{figure*}[t]
    \centering
    \includegraphics[width=0.82\textwidth]{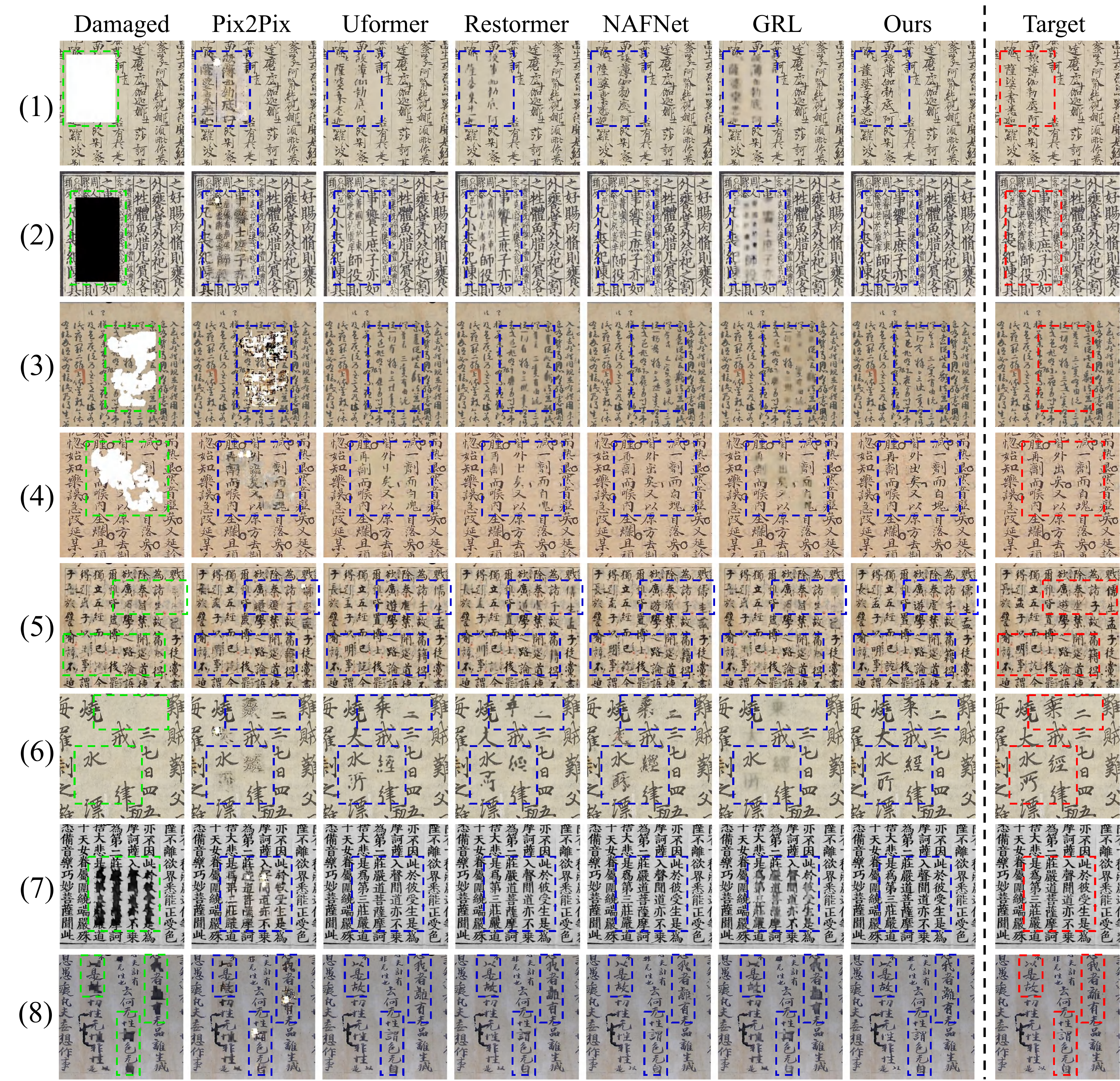}
    \caption{Qualitative comparison. We visualize the results of some evaluated methods. The green, blue, and red boxes represent the damaged regions, the repaired regions, and the target, respectively.}
    \label{fig:qualitative_comparison}
\end{figure*}

\subsubsection{Qualitative Comparison}
As illustrated in Figure \ref{fig:qualitative_comparison}, we present the visualizations of DiffHDR and the existing methods on HDR28K testing set. 
DiffHDR can reconstruct the original appearance of damaged images with both realism and high quality. 
The second-best NAFNet and the third-best Uformer-Big encounter the problems such as blurring (see Figure \ref{fig:qualitative_comparison}(1)(3)(5)), missing character strokes (see Figure \ref{fig:qualitative_comparison}(1)(2)(4)(6)) and style inconsistency of background (see Figure \ref{fig:qualitative_comparison}(3)(7)) within the repaired regions. 
In contrast, DiffHDR excels in these aspects and shows
the superiority on the generation of scribble characters (see Figure \ref{fig:qualitative_comparison}(1)(3)(5)(8)), complex characters (see Figure \ref{fig:qualitative_comparison}(2)(4)(7)), intensive text (see Figure \ref{fig:qualitative_comparison}(1)(2)(3)), and complex background (see Figure \ref{fig:qualitative_comparison}(4)(5)) within repaired regions.

\subsection{Real Damaged Document Image Repair}
\begin{figure*}[t]
    \centering
    \includegraphics[width=\textwidth]{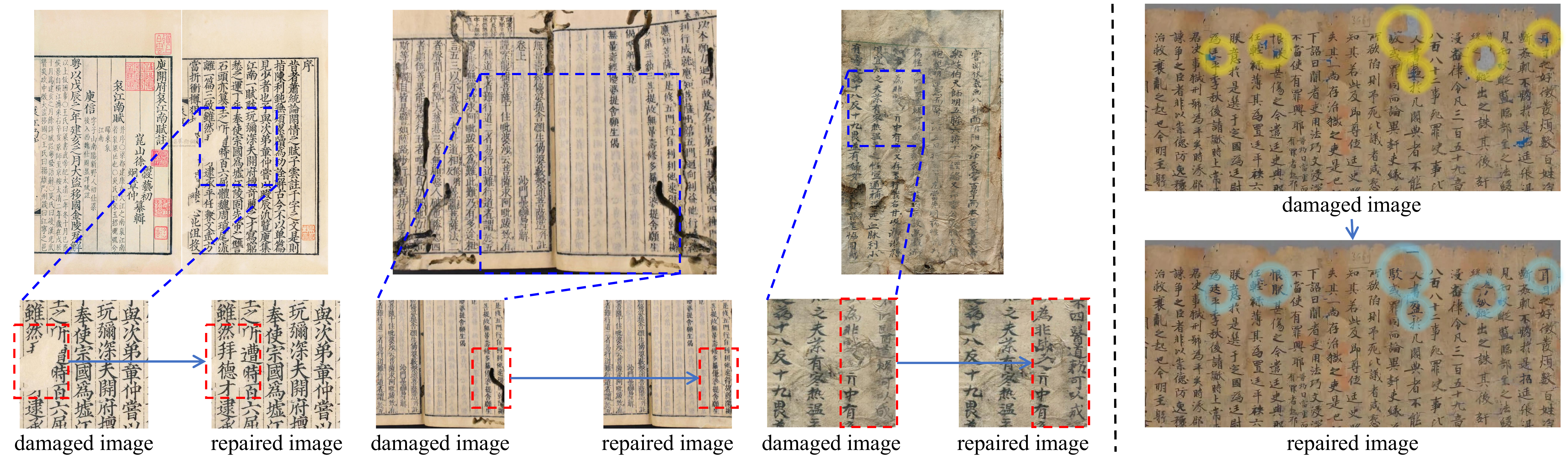}
    \caption{Real damaged historical documents repair by DiffHDR.}
    \label{fig:ablation_unseen}
\end{figure*}

In this section, we utilize the trained DiffHDR to repair real damaged historical document images that are obtained from the Internet.
As shown in Figure \ref{fig:ablation_unseen}, DiffHDR is capable of generating realistic characters coherent with the background during the repair, demonstrating the adaptability of our method in real-world scenarios.
Additionally, it validates the appropriateness of our HDR28K dataset though it is constructed through synthetic degradations.
Note that real damaged-repair image pairs in historical documents are exceptionally rare, making it challenging to collect sufficient data for evaluation purposes. 
In the future, we intend to work with relevant restoration institutions to acquire real damaged-repair pairs to address this issue.

\subsection{Effectiveness of CPLoss $\mathcal{L}_{CP}$}
We investigate the advantage of the proposed Character Content Perceptual Loss $\mathcal{L}_{CP}$, in which we trained the DiffHDR with and without $\mathcal{L}_{CP}$. 
As shown in Figure \ref{tab:cploss}, incorporating the CPLoss improves the repair performance in terms of FID, LPIPS, and  Rec-ACC.

\begin{table}[]
    \centering
    \resizebox{0.73\columnwidth}{!}{%
    \begin{tabular}{@{}lccc@{}}
    \toprule
    \multicolumn{1}{c}{Method} & FID$\downarrow$    & LPIPS$\downarrow$  & Rec-ACC(\%)$\uparrow$ \\ \midrule
    wo $\mathcal{L}_{CP}$                       & 0.8416 & 0.0450 & 64.4897  \\
    w $\mathcal{L}_{CP}$                        & \textbf{0.7499} & \textbf{0.0384} & \textbf{81.9180}  \\ \bottomrule
    \end{tabular}%
    }
    \caption{Effectiveness of character perceptual loss $\mathcal{L}_{CP}$.}
    \label{tab:cploss}
\end{table}

\subsection{Editing and Text Block Font Generation}
% Details and visualizations will be presented in the subsequent sections.
In this section, we explore the capabilities of DiffHDR in historical document editing and text block font generation.
\textbf{(1)} Document editing is to modify the text content to our target while maintaining consistency of style in the edited characters and the surrounding background. 
In our method, given the edited location $\boldsymbol{x}_{m}$ and content $\boldsymbol{x}_c$, we mask the input image by $\boldsymbol{x}_m$ and feed it to DiffHDR. As shown in Figure \ref{fig:other_application}(a), our method generates readable characters, which are also harmonized with the surrounding background.
\textbf{(2)} Text block font generation is to generate a group of characters within the specified region, while the text block adopts the style of the remaining areas. 
As shown in Figure \ref{fig:other_application}(b), DiffHDR generates characters are coherent with the background, though the background area is filled with noise.

\begin{figure}
    \centering
    \includegraphics[width=1\columnwidth]{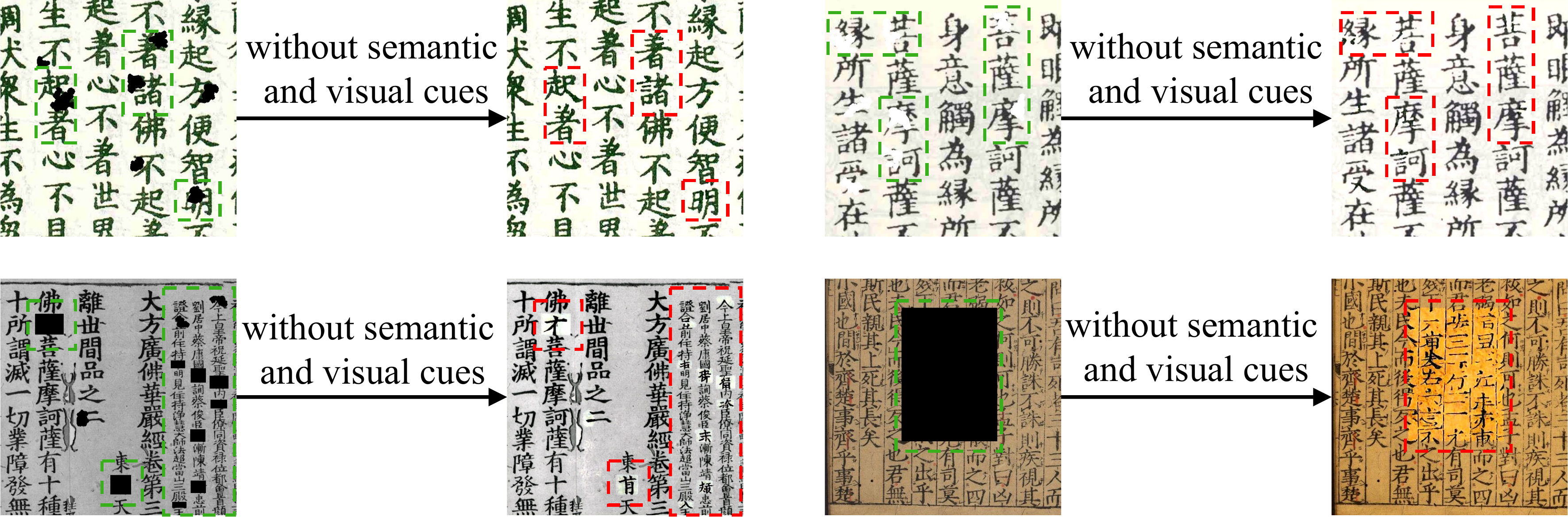}
    \caption{Damaged historical documents repair by DiffHDR when not provided with semantic and spatial cues.}
    \label{fig:semantic_comprehension}
\end{figure}

\subsection{Limitation}
As depicted in the 1\textit{st} row of Figure \ref{fig:semantic_comprehension}, when DiffHDR is not provided with the semantic and spatial information of damaged characters (setting $\boldsymbol{x}_c$ and $\boldsymbol{x}_m$ to pixel $255$), our method can comprehend the character semantics and repair the characters correctly. 
However, when the damage is severe, our method is unable to repair the image in the absence of semantic and spatial information, as shown in the 2\textit{nd} row of Figure~\ref{fig:semantic_comprehension}. 
% For future work, we aim to address this limitation by using the large language model for obtaining semantic information and the layout generation model for spatial cues.
For future work, we will address the prediction of damaged character content and location. Specifically, we plan to leverage large vision-language models (such as Qwen2-VL\cite{wang2024qwen2} and InternVL2\cite{chen2024internvl}) for automated content prediction of damaged characters and to train a detection model (such as the DINO\cite{zhang2022dino}) for identifying damaged character locations.
Moreover, we will collect a certain amount of real damaged-repair image pairs to better evaluate the repair performance of real damaged documents using different methods.

\section{Conclusion}
In this paper, we introduce a new task, \textit{Historical Document Repair} (HDR), which aims to predict the original appearance of damaged historical documents. 
To fill the blank in this field, we contribute a large-scale HDR dataset, named HDR28K, which contains 28,552 damaged-repaired document image pairs and employs three meticulously designed synthetic degradations to simulate real damages typically observed in historical documents. 
Furthermore, a novel DiffHDR model is proposed to solve the HDR problem.
Specifically, DiffHDR follows a diffusion-based paradigm conditioned on semantic and spatial priors for context correctness and visual truthfulness. 
During training, a new character perceptual loss is incorporated to enhance the content preservation of repaired characters. 
Extensive experiments demonstrate that DiffHDR achieves state-of-the-art performance and is capable of repairing real damaged documents though trained with synthetic damages of HDR28K.
Thanks to its highly flexible framework, DiffHDR also exhibits impressive performance in document editing and text block generation.
We believe this study could be the cornerstone of the new HDR field and significantly contribute to the preservation of invaluable cultural heritage.

\section{Acknowledgments}
This research is supported in part by the National Natural Science Foundation of China (Grant No.: 62441604, 62476093) and IntSig-SCUT Joint Lab Foundation.

\bibliography{aaai25}

\clearpage
\section{Diversity of HDR28K}
\begin{figure}[h]
    \centering
    \includegraphics[width=\columnwidth]{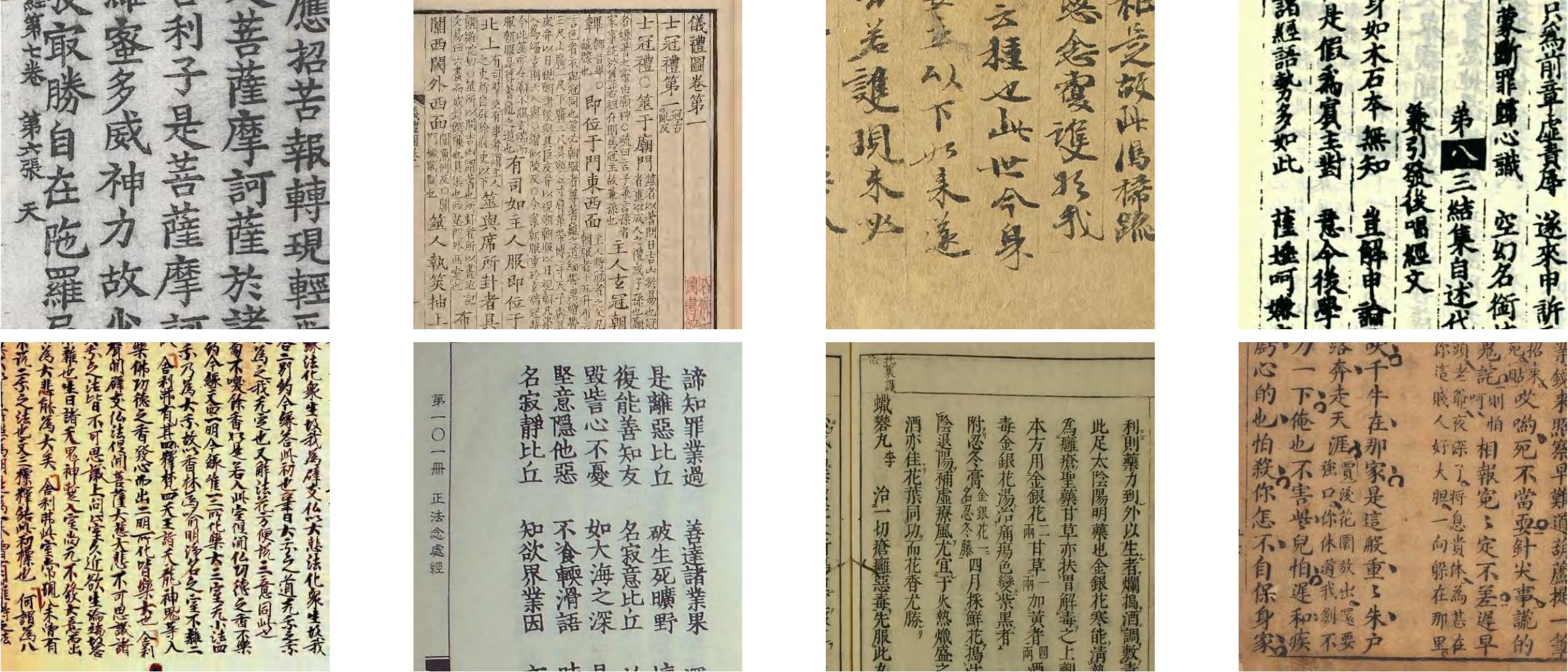}
    \caption{Examples of diverse backgrounds in HDR28K.}
    \label{fig:background}
\end{figure}

\begin{figure}[h]
    \centering
    \includegraphics[width=\columnwidth]{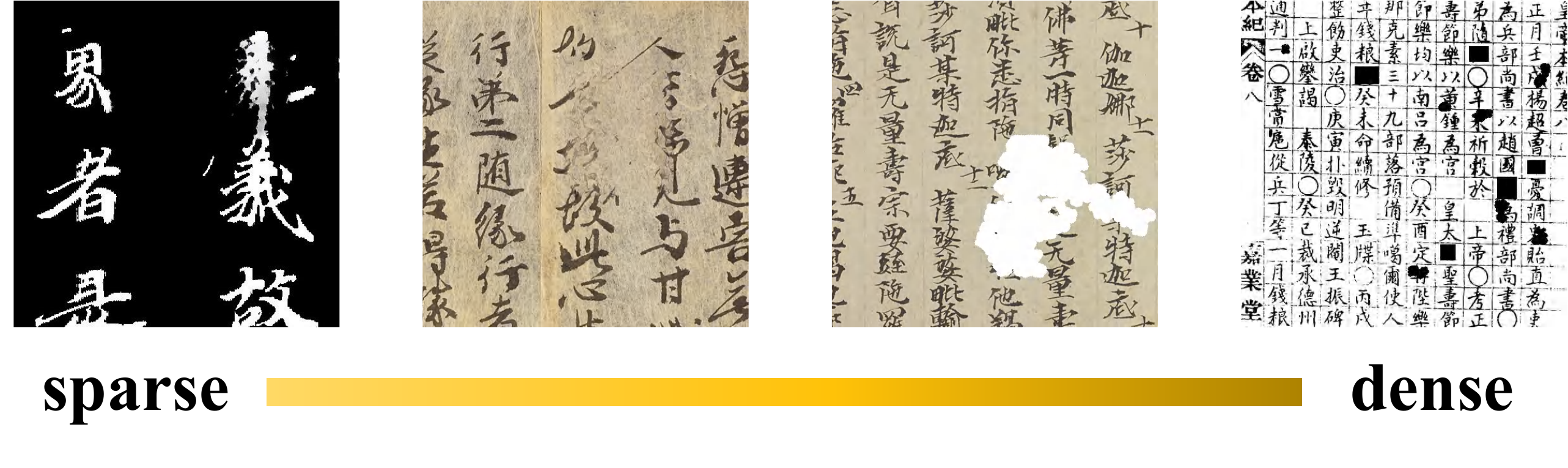}
    \caption{Examples of diverse text densities in HDR28K. Text density is increasing from left to right.}
    \label{fig:text_density}
\end{figure}

\begin{figure}[h]
    \centering
    \includegraphics[width=\columnwidth]{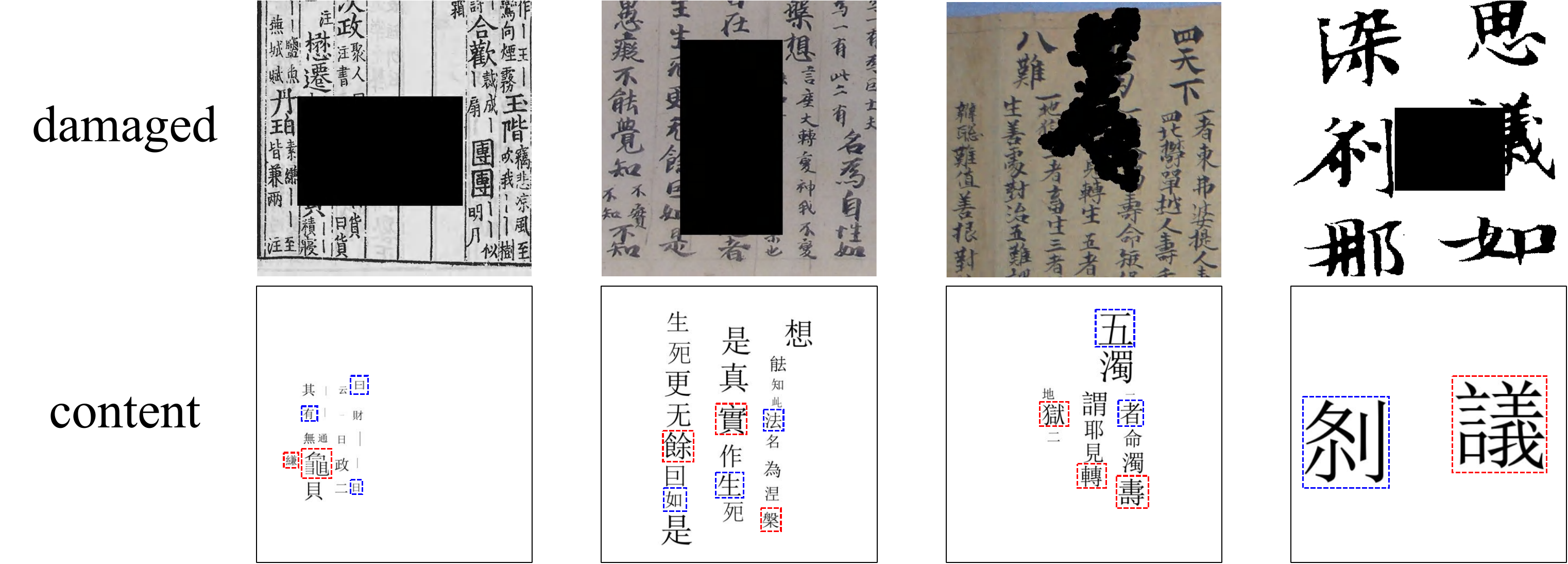}
    \caption{Examples of diverse character complexities in HDR28K. Red boxes denote the characters of hard complexity and blue boxes represent the characters of easy complexity.}
    \label{fig:chareacter_complexity}
\end{figure}

\begin{figure}[h]
    \centering
    \includegraphics[width=\columnwidth]{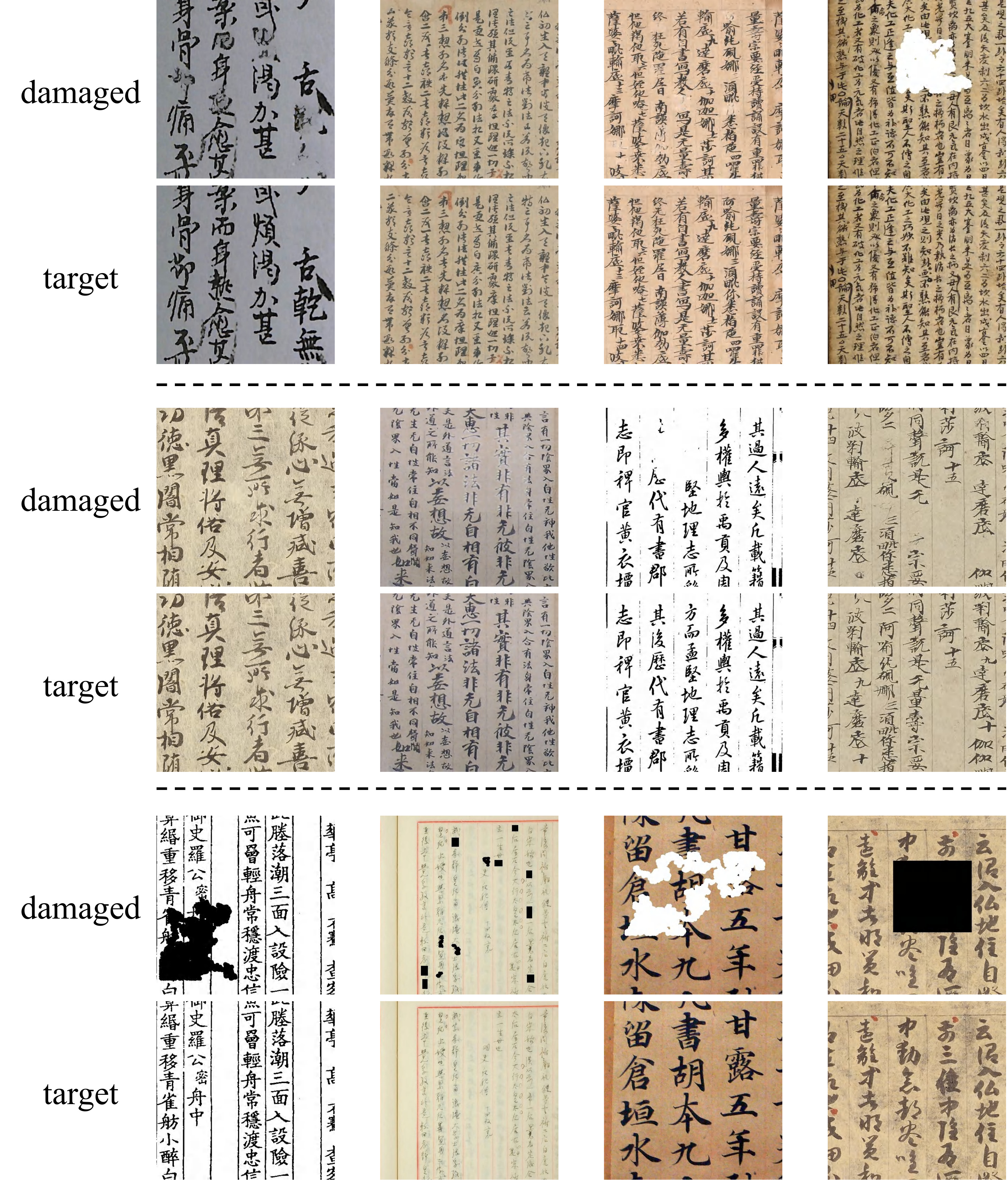}
    \caption{Examples of diverse character styles in HDR28K.}
    \label{fig:character_styles}
\end{figure}

In the HDR28K dataset, multi-style degradations are applied to generate damaged images, thereby enhancing the diversity of damage types. 
Notably, as illustrated from Figures \ref{fig:background} to \ref{fig:character_styles}, HDR28K also exhibits diversity in document backgrounds, text densities, character complexities, and character styles. 
Thus, our proposed dataset stands as highly representative and holds substantial value for the advancement of techniques related to historical document repair.

\section{Repair results on HDR28K using DiffHDR}
\begin{figure*}[h]
    \centering
    \includegraphics[width=\textwidth]{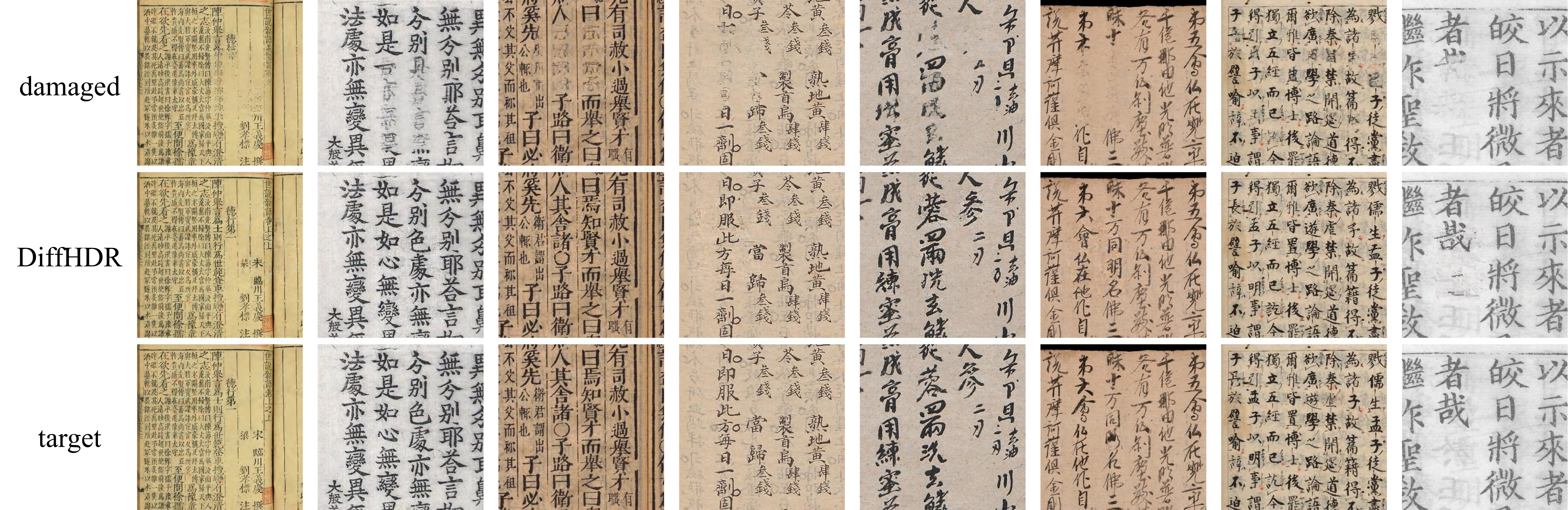}
    \caption{Historical document repair on the type of character missing in HDR28K}
    \label{fig:character_missing}
\end{figure*}

\begin{figure*}[h]
    \centering
    \includegraphics[width=\textwidth]{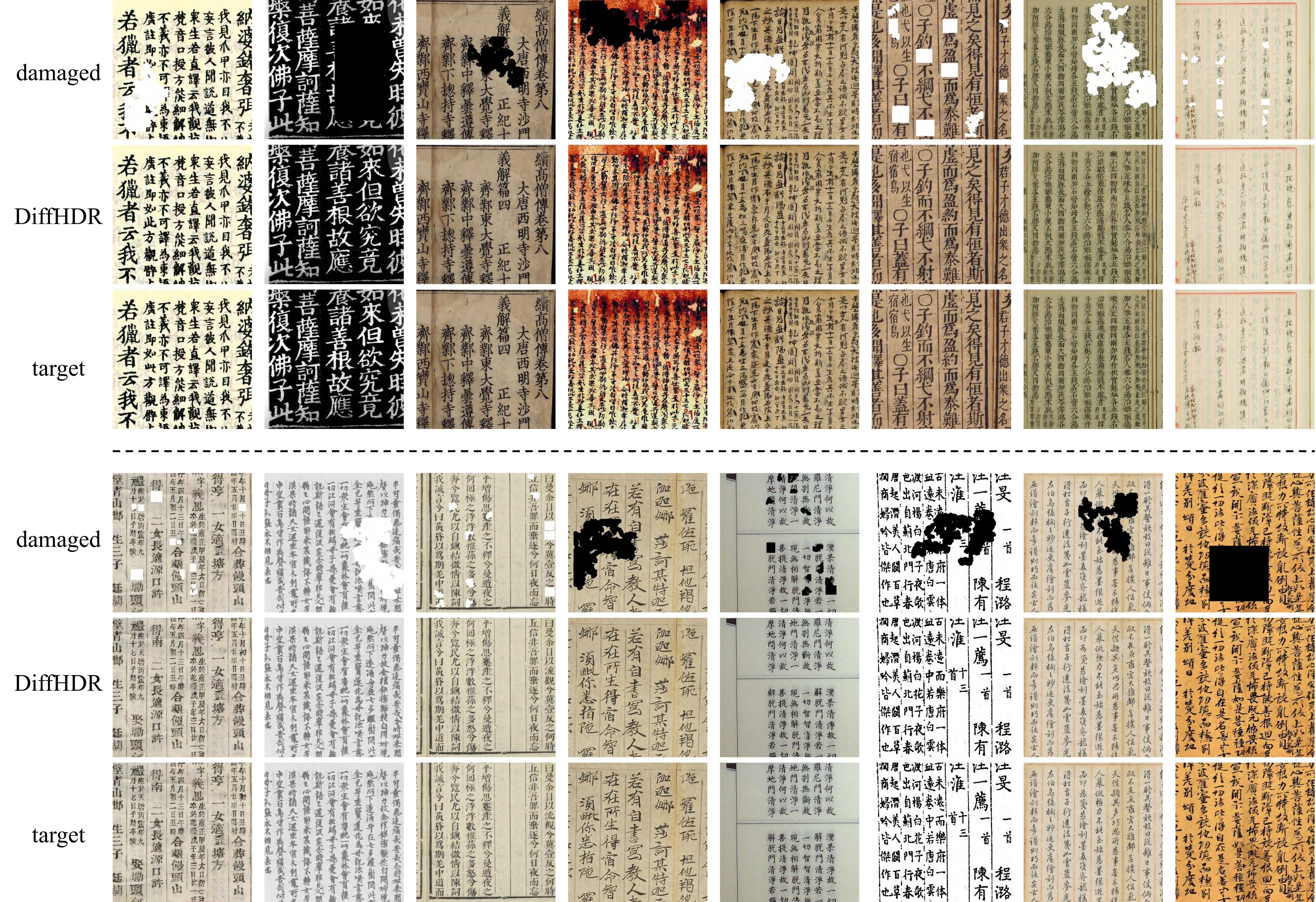}
    \caption{Historical document repair on the type of paper damage in HDR28K}
    \label{fig:paper_damage}
\end{figure*}

\begin{figure*}[h]
    \centering
    \includegraphics[width=\textwidth]{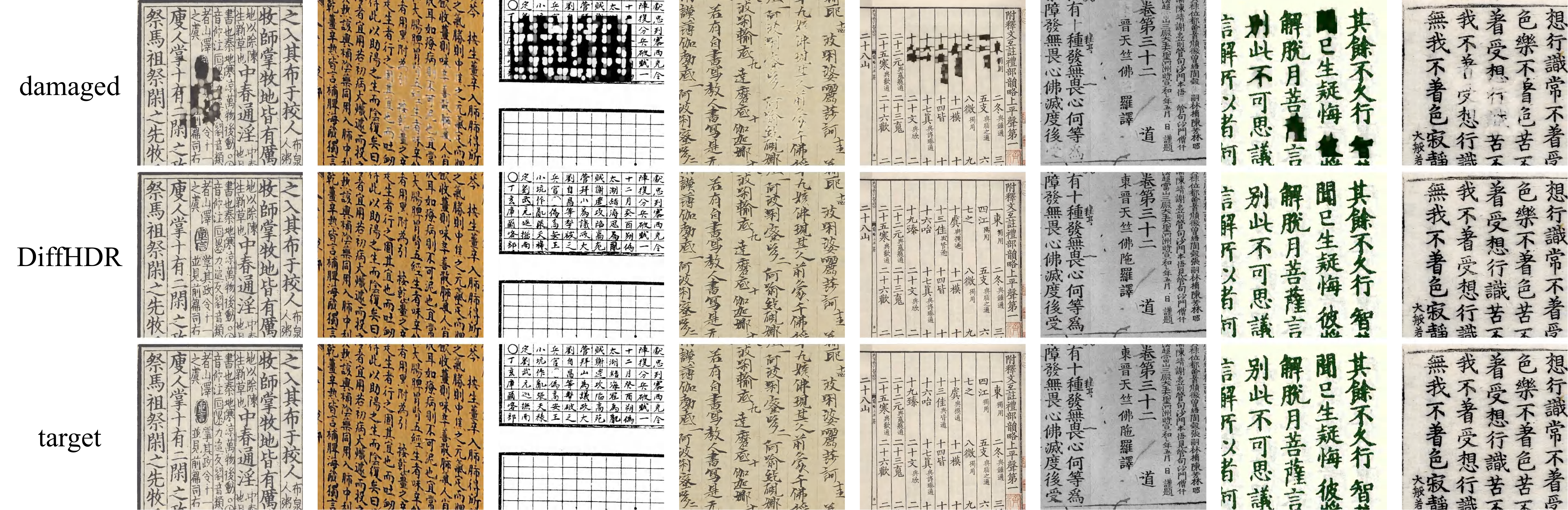}
    \caption{Historical document repair on the type of ink erosion in HDR28K}
    \label{fig:ink_erosion}
\end{figure*}

As shown from Figure \ref{fig:character_missing} to \ref{fig:ink_erosion}, we provide the repair results using DiffHDR on the samples of character missing, paper damaged, and ink erosion, respectively.
It demonstrates that DiffHDR excels in repairing the historical documents of the three degraded types and can be capable of handling complex backgrounds, intricate character styles (such as scribble style), and text of diverse densities.

\end{document}